\documentclass[letterpaper]{article} 
\usepackage{aaai24}  
\usepackage{times}  
\usepackage{helvet}  
\usepackage{courier}  
\usepackage[hyphens]{url}  
\usepackage{graphicx} 
\usepackage{natbib}  
\usepackage{caption} 
\usepackage{algorithm}
\usepackage{algorithmic}
\usepackage{multirow}
\usepackage{bm}
\usepackage{amsmath}
\usepackage{float}
\usepackage{amssymb}

\usepackage{newfloat}
\usepackage{listings}
\DeclareCaptionStyle{ruled}{labelfont=normalfont,labelsep=colon,strut=off} 
\floatstyle{ruled}
\newfloat{listing}{tb}{lst}{}
\floatname{listing}{Listing}
\usepackage{adjustbox}

\title{ST-MLP: A Cascaded Spatio-Temporal Linear Framework with Channel-Independence Strategy for Traffic Forecasting}

\author {
    Zepu Wang\equalcontrib\textsuperscript{\rm 1},
    Yuqi Nie\equalcontrib\textsuperscript{\rm 2},
    Peng Sun\textsuperscript{\rm 3},
    Nam H. Nguyen\textsuperscript{\rm 4},
    John Mulvey \textsuperscript{\rm 2},
    H. Vincent Poor \textsuperscript{\rm 2} 
}
\affiliations {
    \textsuperscript{\rm 1}University of Pennsylvania
    \textsuperscript{\rm 2} Princeton University 
    \textsuperscript{\rm 3} Duke Kunshan University 
    \textsuperscript{\rm 4} IBM Research \\
     zepu@seas.upenn.edu, 
     peng.sun568@duke.edu,
    nnguyen@us.ibm.com,
    \{ynie,mulvey,poor\}@princeton.edu
}

\usepackage{bibentry}
\usepackage{textcomp}
\usepackage{xcolor}

\begin{document}

\maketitle

\begin{abstract}

The criticality of prompt and precise traffic forecasting in optimising traffic flow management in Intelligent Transportation Systems (ITS) has drawn substantial scholarly focus. Spatio-Temporal Graph Neural Networks (STGNNs) have been lauded for their adaptability to road graph structures. Yet, current research on STGNNs architectures often prioritises complex designs, leading to elevated computational burdens with only minor enhancements in accuracy. To address this issue, we propose ST-MLP, a concise spatio-temporal model solely based on cascaded Multi-Layer Perceptron (MLP) modules and linear layers. Specifically, we incorporate temporal information, spatial information and predefined graph structure with a successful implementation of the channel-independence strategy - an effective technique in time series forecasting. Empirical results demonstrate that ST-MLP outperforms state-of-the-art STGNNs and other models in terms of accuracy and computational efficiency. Our finding encourages further exploration of more concise and effective neural network architectures in the field of traffic forecasting.

\end{abstract}

\section{Introduction}

Traffic forecasting is pivotal in the Intelligent Traffic System (ITS) since it can provide accurate road status data that aids informed decision-making for agencies~\cite{wang2023st, wang2022novel, wang2022novel1}. Consequently, it has gained considerable attention from both academia and industry~\cite{wang2023novel, wang2022sfl}.

Traffic forecasting is often treated as a spatio-temporal time series forecasting problem~\cite{jin2023survey}. Thanks to the rise of deep learning, spatio-temporal graph neural networks (STGNNs) have become a popular research direction. Various STGNNs have been introduced, showing impressive forecasting accuracy~\cite{weng2023decomposition, guo2021learning, guo2019attention}. These models use different strategies: some rely on predefined adjacency matrices to capture spatial correlations~\cite{guo2019attention, wu2019graph}, while others use neural networks to extract dynamic spatial patterns~\cite{guo2021learning, bai2020adaptive, oreshkin2021fc}. A few even combine both methods~\cite{shao2022decoupled}. However, recent STGNNs have grown more complex without delivering much better performance, leading to longer computation times and higher memory usage~\cite{shao2022spatial}. This situation has prompted a reevaluation of these complex architectures, sparking a search for simpler and more efficient techniques~\cite{liu2023we}.

In time series forecasting, some simpler yet effective methods like TSMixer~\cite{vijay2023tsmixer} and TiDE~\cite{das2023long} have shown strong performance on different forecasting tasks. In the area of analyzing spatio-temporal data, a straightforward approach called STID by \textit{Shao et al.} combines spatial and temporal information using MLPs, achieving good results while being efficient~\cite{shao2022spatial}. Also, \textit{Qin et al.} designed an architecture based on MLP transformers~\cite{qin2023mlp} that outperforms many complex models. Yet, despite these successful examples, there's still a lack of exploration of concise models in traffic and spatio-temporal forecasting.

\begin{figure}[t]
\begin{center}

\includegraphics[width = \linewidth]{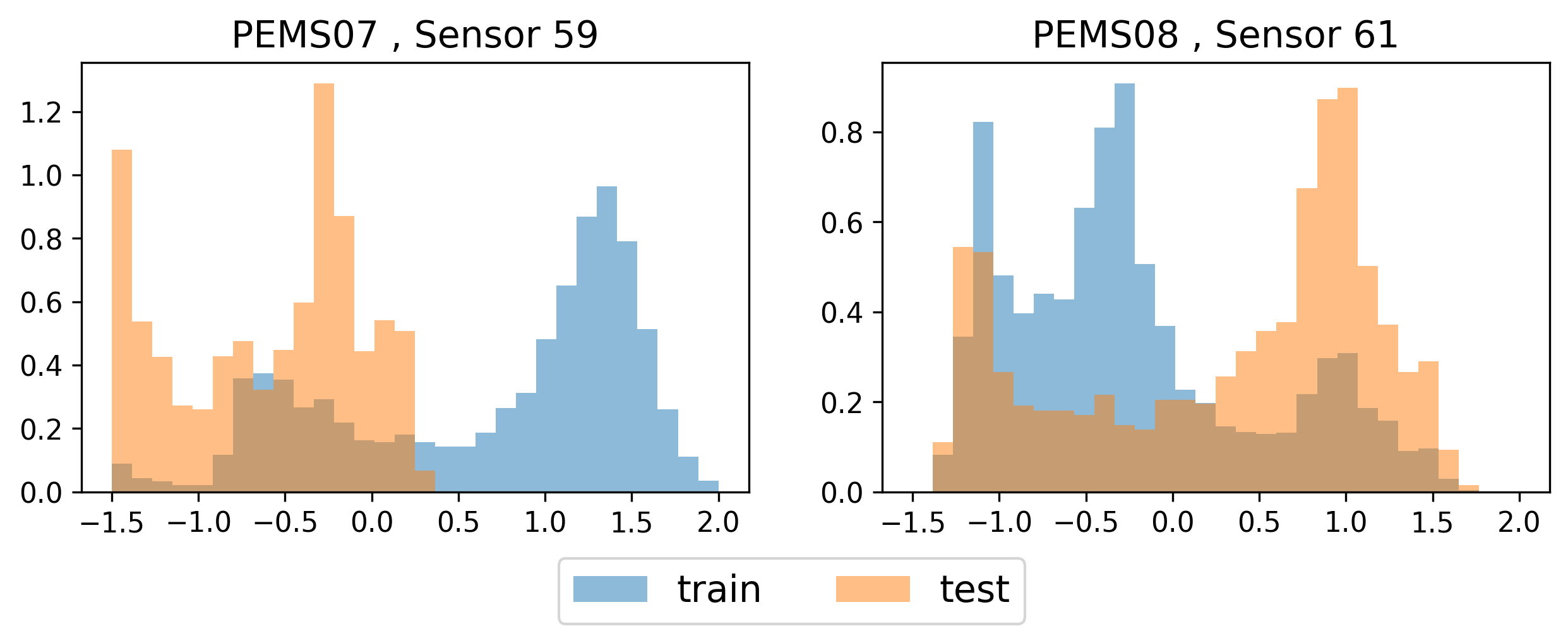}
\end{center}
\caption{Examples of train \& test data distribution shift on PEMS07 and PEMS08 traffic datasets. Data is processed by Z-Score normalization.}
\label{fig:distribution}
\end{figure}

Moreover, dealing with the issue of distribution shift between training and testing data has been a common challenge in multivariate time series forecasting~\cite{kim2021revin}. This shift can have negative impacts, including problems related to overfitting. It's important to highlight that this distribution shift problem also emerges in various spatio-temporal forecasting tasks, even after datasets have been normalized, as shown in Figure~\ref{fig:distribution}. Recent research has introduced a technique called "channel-independence" (CI) to tackle distribution shift and mitigate overfitting in long-term time series forecasting~\cite{han2023ci, nie2022time}. To elaborate, CI involves using only the historical data within a single channel to predict future time steps within that same channel, without directly incorporating any information from other channels. In the context of spatio-temporal forecasting tasks, CI can be understood as predicting the future data of a specific node without depending on data from other nodes.

However, CI has not found widespread application in spatio-temporal data analysis, particularly when prior predefined graph structures are involved. Many models tend to capture spatio-temporal correlations within data by employing various techniques along both the temporal and spatial dimensions, often leading to channel-mixing (CM). CM stands in contrast to the concept of CI. This prompts us to think: is it feasible to harness the benefits of CI while simultaneously retaining the ability to capture the underlying spatio-temporal relationships within traffic patterns?



In this paper, we propose a spatio-temporal MLP (ST-MLP), a cascaded channel-independent framework for efficient traffic forecasting. The main contributions of our work can be summarized as follows:

\begin{itemize}
\item[$\bullet$] We propose a concise structure that is solely based on MLPs, ensuring its simplicity and efficiency.
\item[$\bullet$] To our best knowledge, our work is the first one to apply the channel-independence strategy in spatio-temporal forecasting problems while incorporating information from predefined graphs.
\item[$\bullet$] In an effort to enhance the integration of diverse embeddings, we implement a cascaded structure. Comparative analysis demonstrates that this method outperforms the conventional approach of simple concatenation.
\item[$\bullet$] Extensive evaluation on various traffic datasets show the superiority of ST-MLP by contrasting its results against those of more than 10 state-of-the-art models.




\end{itemize}


\section{Preliminary}
\label{Problem Statement}

\subsection{Problem Definition}

Formally, a graph $G$ is defined as an ordered pair $(V,E)$, where $V$ represents the set of vertices (or nodes) and $E$ represents the set of edges. In this paper, we construct the adjacency matrix $A$ of $G$ by scaling the normalized Laplacian matrix~\cite{kipf2016semi}. Suppose the historical traffic data is embedded on graph $G$ with $C$ multivariate traffic features and $T$ time intervals. Thus, the historical time series feature can be denoted as $X \in \mathbb{R}^{N \times T \times C}$.

The objective of traffic forecasting is to learn a mapping function $f$ that takes historical traffic data $X$ and graph $G$ as inputs to predict future traffic data for $Q$ time intervals, denoted as $\hat{X} \in \mathbb{R}^{N \times Q \times C}$. For simplicity, in this paper, we set $C$ = 1, resulting in $X \in \mathbb{R}^{N \times T}$ and $\hat{X} \in \mathbb{R}^{N \times Q}$. Subsequently, we refer to the first dimension as the channel dimension and the second dimension as the temporal dimension.

Drawing on insights from ~\textit{Chen et al.}, we can enhance the forecasting performance by incorporating additional time-related information. Our focus will be specifically on Time in Day (TD) and Day in Week (DW). TD represents the index of a particular time slot within a day, while DW signifies the weekday label. For any given input data $X$, the corresponding TD and DW can be represented as two integral vectors: $T_{td} \in \mathbb{R}^{T}$ and $T_{dw} \in \mathbb{R}^{T}$, respectively, with these vectors remaining consistent across all nodes. The concepts of TD and DW are directly tied to the recurring daily and weekly patterns of traffic flow due to the evident periodic nature of traffic.


Therefore, the forecast task could be expressed as:

\begin{equation}
\hat{X} = f(G, X, T_{td}, T_{dw}).
\end{equation}

\subsection{Channel-independence on Spatio-temporal Data}

A multivariate time series can be analogously seen as a multi-channel signal. In the context of spatio-temporal forecasting tasks, we designate the spatial dimension as channels, where each of the $N$ nodes corresponds to a distinct channel. Assuming we have an input data denoted as $X_{input} \in \mathbb{R}^{N \times T}$ for processing, there exist three fundamental types of modules:

\begin{gather}
    X_t = \texttt{TemporalMix} (X_{input}), \label{tmixing}\\
    X_c = \texttt{ChannelMix} (X_{input}), \label{cmixing}\\
    X_{tc} = \texttt{TemporalChannelMix} (X_{input}). \label{tcmixing}
\end{gather}

In this context, the \texttt{TemporalMix} module maps the temporal dimension of $X_{input}$ into $X_t \in \mathbb{R}^{N \times T'}$; the \texttt{ChannelMix} module, conversely, maps the channel (spatial) dimension of $X_{input}$ into $X_t \in \mathbb{R}^{N' \times T}$; and the \texttt{TemporalChannelMix} module operates both dimensions, yielding an output denoted as $X_{tc} \in \mathbb{R}^{N' \times T'}$. Here, $T'$ signifies the intended temporal dimension, while $N'$ represents the intended spatial dimension for the output of these three core modules. It is noteworthy that models adopting the CI strategy incorporate solely modules as described in Equation (\ref{tmixing}), whereas CM models have the flexibility to encompass any of these module types.

The CI approach serves a dual purpose: it not only addresses the challenge of distribution shift but also leads to a reduction in the variance of prediction outcomes~\cite{han2023ci}. To preserve the inherent spatio-temporal correlations while reaping the advantages of CI, this paper adopts a strategy wherein all temporal and spatial information, inclusive of the underlying graph structure, is amalgamated within the temporal dimension. This arrangement facilitates the autonomous prediction of each channel while retaining the benefits of independent forecasting.

\section{Proposed Method}
\label{Proposed Method}

In this section, we would like to explain the general ST-MLP structure and its components in details.

\begin{figure*}[t]
\begin{center}
\includegraphics[width= 0.9\linewidth]
{ 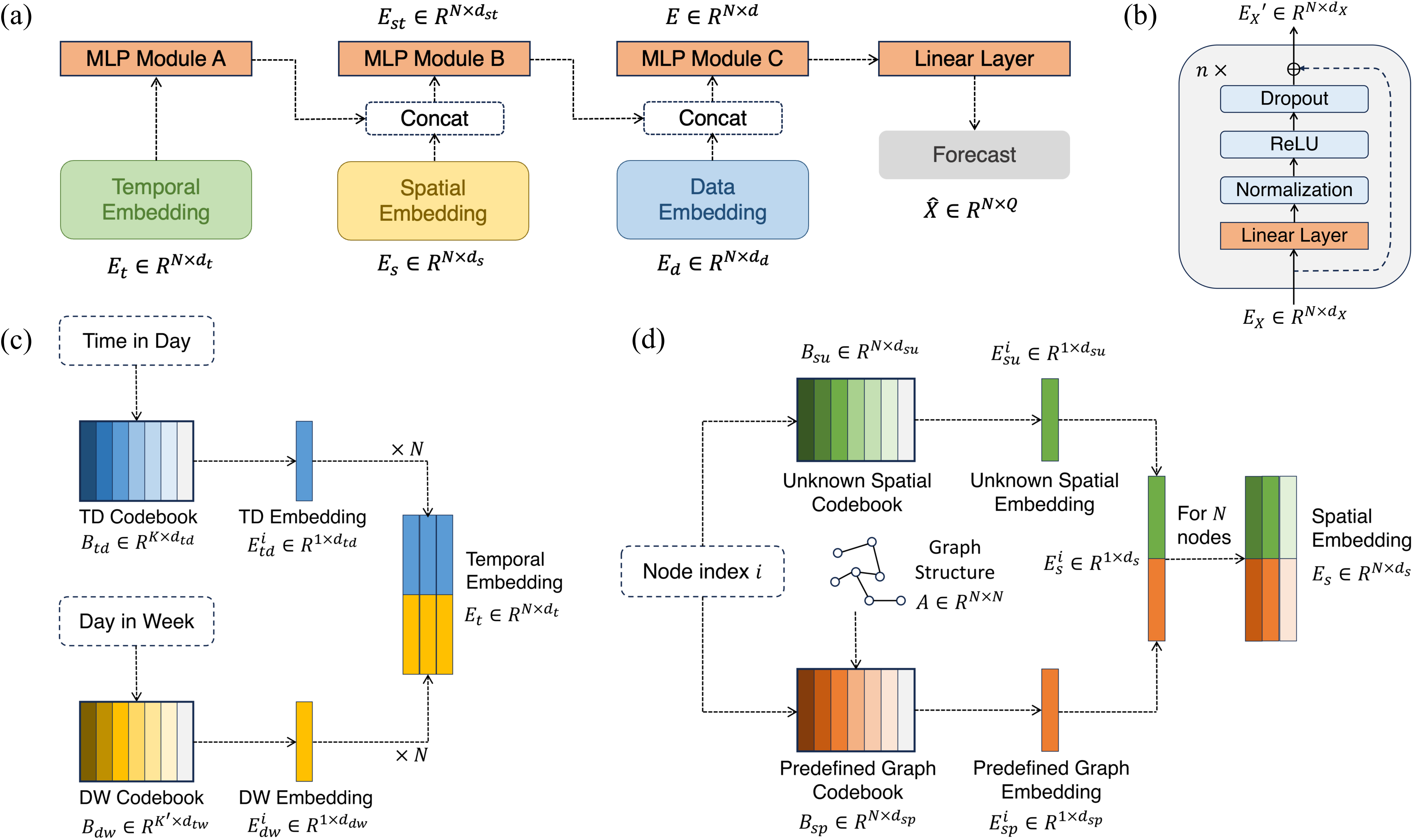}
\end{center}
\caption{(a) Architecture of ST-MLP. It uses temporal embedding, spatial embedding and data embedding to generate the forecast result. (b) Structure of MLP Module in Figure~\ref{fig:model}(a). (c) Details of Temporal Embedding, which contains information from Time in Day and Day in Week. (d) Details of Spatial Embedding, which consists of unknown spatial embedding and predefined graph embedding.}
\label{fig:model}
\end{figure*}

\subsection{General Structure}

As illustrated in Figure~\ref{fig:model} (a), ST-MLP adopts a cascaded structure to merge different types of information in distinct embeddings. In our study, the symbols marked as $d_X$ stand for hyperparameters used to define the temporal dimension of various embeddings. Here, the symbol $X$ acts as a variable that adjusts to different scenarios. Further details about $d_X$ and all other symbols are outlined in the Appendix. 
Specifically, our model takes into account three crucial embeddings: temporal embedding $E_{t} \in \mathbb{R}^{N \times d_{t}}$, which holds crucial temporal details of the traffic data; spatial embedding $E_{s} \in \mathbb{R}^{N \times d_{s}}$, representing the spatial correlations in the traffic graph; and data embedding $E_{d} \in \mathbb{R}^{N \times d_d}$. The specifics of each embedding will be explained in the upcoming sections.

In the initial stage, the temporal embedding undergoes processing through the MLP module A. Following this, it gets combined with the spatial embedding in the subsequent stage. This combined data then undergoes processing by the MLP module B, resulting in the creation of a spatio-temporal embedding denoted as $E_{st} \in \mathbb{R}^{N \times d_{st}}$, where $d_{st} = d_{s} + d_{t}$. Moving on to the third stage, the spatio-temporal embedding is merged with the data embedding. Here, the MLP module C is applied, generating a comprehensive embedding $E \in \mathbb{R}^{N \times d}$, where $d = d_{st} + d_d$. In the final step, a linear layer is employed to map $E$ to the forecasted outcome $\hat{X} \in \mathbb{R}^{N \times Q}$. It's important to note that all these operations are carried out within the temporal dimension to ensure the application of the CI strategy.

\subsection{MLP Module}

In Figure~\ref{fig:model} (b), the general and straightforward design of the MLP module is depicted. In the cases of modules A, B, and C, we repeat the basic block $n_A$, $n_B$, and $n_C$ times respectively. These values are adjustable hyperparameters. Each basic block encompasses a linear layer that solely operates on the embedding dimension $d_X$, a normalization layer for stability during training, a ReLU activation function, a dropout layer to curb overfitting, and a residual connection. It's worth noting that we maintain the CI strategy within the MLP modules since we are only operating the temporal dimension.

\subsection{Temporal Embedding}

\subsubsection{Embedding of Time in Day}

Let's consider a day divided into $K$ time slots equally. When we're given $T_{td}$, the final time index in $T_{td}$ is used. This index is then transformed into a one-hot vector, $t_{td} \in \mathbb{R}^{1 \times K}$, which signifies the Time in Day (TD) index for the entire sequence. Alongside this, there's a codebook $B_{td} \in \mathbb{R}^{K \times d_{td}}$ that can be learned. In Figure \ref{fig:model} (c), when we perform a dot product of $t_{td} \cdot B_{td}$, we're essentially selecting the corresponding embedding vector $E_{td}^i \in \mathbb{R}^{1 \times d_{td}}$ that corresponds to a specific TD index. Since all nodes share the same time information, we can replicate $E_{td}^i$ $N$ times to obtain the embedding $E_{td} \in \mathbb{R}^{N \times d_{td}}$, representing the TD details for all nodes.

\subsubsection{Embedding of Day in Week}

Considering there are $K'=7$ days in a week, we can take the index from the end of the sequence $T_{dw}$. This index is transformed into a one-hot vector $t_{dw} \in \mathbb{R}^{1 \times K'}$ that signifies the Day in Week (DW) index for the entire sequence. Similar to the TD embedding, the embedding of DW $E_{dw} \in \mathbb{R}^{N \times d_{dw}}$ can be obtained from a learnable codebook $B_{dw} \in \mathbb{R}^{K' \times d_{dw}}$. This helps us capture weekly patterns.


\subsubsection{Temporal Embedding Representation}
As shown in Figure~\ref{fig:model} (c), $E_{td}$ and $E_{dw}$ are generated paralleled, then are concatenated together:
\begin{equation}
E_{t} = \texttt{Concat}[(E_{td},E_{dw}),\texttt{dim} = 1].
\end{equation}
Hence, we can derive the general temporal embedding $E_{t} \in \mathbb{R}^{N \times d_{t}}$, where $d_{t} = d_{td} + d_{dw}$.

\subsection{Spatial Embedding}

\subsubsection{Embedding of known predefined graph structure}

With $A \in \mathbb{R}^{N \times N}$ representing a predefined adjacency matrix, we aim to integrate fixed graph details into the neural network. To achieve this, we start by initializing a learnable codebook $B_{sp} \in \mathbb{R}^{N \times d_{sp}}$. We then utilize matrix multiplication to proceed:
\begin{equation}
A B_{sp} = E_{sp},
\end{equation}
where $E_{sp} \in \mathbb{R}^{N \times d_{sp}}$. Next, we employ the node index $i$ to select the appropriate embedding vector $E_{sp}^i \in \mathbb{R}^{1 \times d_{sp}}$, as illustrated in Figure \ref{fig:model} (d). This approach allows us to include predetermined graph information while retaining the CI principle within the model structure.

\subsubsection{Embedding of unknown spatial information}

Previous studies underscore that a pre-defined graph may not fully encapsulate all the authentic spatial intricacies inherent in real-world scenarios~\cite{zhang2020spatio,guo2021learning}. To uncover hidden spatial details from traffic data, we introduce an additional learnable codebook $B_{su} \in \mathbb{R}^{N \times d_{su}}$ to signify inherent spatial characteristics. Similar to the procedure discussed in the preceding paragraph while learning $E_{sp}^i$, we acquire $E_{su}^i \in \mathbb{R}^{1 \times d_{su}}$ for node $i$.

\subsubsection{Spatial Embedding Representation}

In Figure~\ref{fig:model} (d), similar to the approach used for temporal representation, the overall spatial representation is obtained by concatenating two spatial embeddings:
\begin{equation}
E_{s}^i = \texttt{Concat}[(E_{sp}^i,E_{su}^i),\texttt{dim} = 1].
\end{equation}
By repeating this for all $N$ nodes, we can get the final spatial embedding $E_{s} \in \mathbb{R}^{N \times d_{s}}$, where $d_{s}=d_{sp}+d_{su}$.

\subsection{Data Embedding}

We replicate $T_{td}$ for $N$ times, resulting in $T_{td}'\in \mathbb{R}^{N \times T}$. Similarly, we repeat the procedure with $T_{dw}$ to generate $T_{dw}'\in \mathbb{R}^{N \times T}$. As for the data embedding, we concatenate $X$, $T_{td}'$, and $T_{dw}'$ to construct a representation that encapsulates both the original traffic data and time stamps. Ultimately, a linear transformation is employed to map the temporal dimension to $d_d$:

\begin{equation}
E_d = \texttt{Linear}(\texttt{Concat}[(X, T_{td}', T_{dw}'), \texttt{dim} = 1]),
\end{equation}
where $E_d \in \mathbb{R}^{N \times d_d}$.


\section{Experiments}
\label{EXPERIMENT}
\subsection{Datasets}

We employ four widely used public traffic datasets to conduct a comparative analysis of forecasting performance. These datasets encompass a variety of traffic data types, including traffic speed and volume. Specifically, we utilize the PEMS-BAY dataset for traffic speed and the PEMS04, PEMS07, and PEMS08 datasets for traffic volume, as documented by Chen et al.~\cite{chen2001pems,shao2022basicts}. The traffic measurements in these datasets are collected from loop sensors situated on highway networks. These measurements are aggregated into 5-minute intervals and subsequently subjected to Z-Score normalization. Our data is divided into three subsets: training (70\%), validation (10\%), and testing (20\%).

\subsection{Baseline Methods}

We compare the proposed model (ST-MLP) with the following methods for traffic forecasting. For 
conventional method, we choose Historical Inertia (HI)~\cite{cui2021historical}. For STGNNs Methods, we choose
Graph WaveNet~\cite{wu2019graph}, DCRNN~\cite{li2017diffusion}, AGCRN~\cite{bai2020adaptive},
STGCN~\cite{yu2017spatio},
StemGNN~\cite{cao2020spectral},
GTS~\cite{shang2021discrete},
MTGNN~\cite{wu2020connecting},
DGCRN~\cite{li2023dynamic}. For time series forecasting methods. we choose
DLinear~\cite{zeng2023transformers}, and PatchTST~\cite{nie2022time}. In addition, we also apply two innovative methods:  STID~\cite{shao2022spatial} and STNorm~\cite{providence2022spatial}.

We use BasicTS and Easytorch ~\cite{wang2020easytorch,shao2022basicts} to implement baseline models. To ensure fair and consistent evaluations, all baseline models are assessed under identical environments. Details are provided in ~Appendix. 

\subsection{Implementation Details}

In our traffic forecasting benchmarks, we adopt a historical data approach involving a 12-step strategy. This approach uses the last 12 observations, equivalent to a 1-hour interval, to predict the upcoming 12-step observations, also spanning an hour. We consider time intervals of 3 steps (15 minutes), 6 steps (30 minutes), and 12 steps (60 minutes) between historical inputs and target forecasts. Additionally, we evaluate the average performance over the 12 steps. Given that the datasets' time intervals are 5 minutes each, a single day is divided into 288 time intervals. Consequently, the parameter $K$ for embedding Time in Day is set to 288. The training process for all models takes place on an AMD EPYC 7642 Processor @ 2.30GHz with a single RTX 3090 (24GB) GPU.


In natural language processing tasks, Layer Normalization is often favored over Batch Normalization. However, in time series tasks, this issue is still a subject of debate~\cite{nie2022time}. We consider this as a critical hyperparameter in our experiments. To assess the performance of all methods, we employ three metrics: Mean Absolute Error (MAE), Root Mean Square Error (RMSE), and Mean Absolute Percentage Error (MAPE). All the necessary settings for reproducing our results are available in our code\footnote{will be open-sourced upon publication} and Appendix.




\begin{table*}[!ht]
    \centering
    
    \renewcommand{\arraystretch}{0.95} 
    \begin{adjustbox}{width=0.9\textwidth}
    {\fontsize{9}{11}\selectfont
    \begin{tabular}{|p{0.1cm}|c|ccc|ccc|ccc|ccc|}
    \hline
        \multicolumn{1}{|c|}{Datasets} & \multicolumn{1}{c|}{Methods} & \multicolumn{3}{c|}{15 min} & \multicolumn{3}{c|}{30 min} & \multicolumn{3}{c|}{60 min} & \multicolumn{3}{c|}{Average} \\ \cline{3-14}
        \multicolumn{1}{|c|}{} & \multicolumn{1}{c|}{} & MAE & RMSE & MAPE & MAE & RMSE & MAPE & MAE & RMSE & MAPE & MAE & RMSE & MAPE \\ \hline \hline
        \multirow{10}{*}{\rotatebox{90}{PEMS-BAY}} & HI & 3.06 & 7.05 & 6.85\% & 3.06 & 7.04 & 6.84\% & 3.05 & 7.03 & 6.83\% & 3.05 & 7.05 & 6.84\% \\ \cline{2-14} 
        & GraphWaveNet & \textbf{1.31} & \textbf{2.76} & 2.75\% & 1.66 & 3.77 & 3.76\% & 1.98 & 4.58 & 4.69\% & 1.60 & 3.69 & 3.60\% \\ \cline{2-14}
        & DCRNN & \textbf{1.31} & \textbf{2.76} & \textbf{2.73\%} & 1.65 & 3.75 & 3.70\% & 1.97 & 4.58 & 4.66\% & 1.59 & 3.68 & 3.57\% \\ \cline{2-14}
        & AGCRN & 1.35 & 2.83 & 2.91\% & 1.66 & 3.76 & 3.80\% & 1.93 & 4.46 & 4.57\% & 1.60 & 3.65 & 3.64\% \\ \cline{2-14}
        & STGCN & 1.36 & 2.82 & 2.87\% & 1.70 & 3.82 & 3.86\% & 2.01 & 4.59 & 4.73\% & 1.64 & 3.73 & 3.70\% \\ \cline{2-14}
        & StemGNN & 1.57 & 3.26 & 3.45\% & 2.12 & 4.67 & 4.95\% & 2.80 & 6.16 & 6.78\% & 2.09 & 4.72 & 4.89\% \\ \cline{2-14}
        & GTS & 1.36 & 2.87 & 2.85\% & 1.72 & 3.84 & 3.88\% & 2.05 & 4.60 & 4.87\% & 1.66 & 3.75 & 3.73\% \\ \cline{2-14}
        & MTGNN & 1.33 & 2.80 & 2.78\% & 1.64 & 3.73 & 3.71\% & 1.93 & 4.48 & 4.60\% & 1.58 & 3.63 & 3.56\% \\ \cline{2-14}
        & DGCRN & 1.33 & \textbf{2.76} & 2.74\% & 1.65 & 3.74 & \textbf{3.67\%} & 1.98 & 4.58 & 4.62\% & 1.62 & 3.68 & 3.58\% \\ \cline{2-14}
        & DLinear & 1.58 & 3.41 & 3.28\% & 2.15 & 4.89 & 4.71\% & 2.97 & 6.76 & 6.89\% & 2.14 & 5.04 & 4.72\% \\ \cline{2-14}
        & PatchTST & 1.55 & 3.38 & 3.26\% & 2.12 & 4.92 & 4.67\% & 2.98 & 6.93 & 6.88\% & 2.12 & 5.10 & 4.71\% \\ \cline{2-14}
        & STID & \textbf{1.31} & 2.78 & 2.74\% & 1.63 & 3.70 & 3.68\% & \textbf{1.90} & 4.39 & 4.48\% & \textbf{1.56} & 3.60 & 3.51\% \\ \cline{2-14}
        & STNorm & 1.33 & 2.82 & 2.78\% & 1.65 & 3.78 & 3.68\% & 1.91 & 4.45 & 4.47\% & 1.58 & 3.66 & 3.51\% \\ \cline{2-14}
        & ST-MLP & 1.32 & 2.78 & 2.77\% & \textbf{1.62} & \textbf{3.65} & \textbf{3.67\%} & \textbf{1.90} & \textbf{4.34} & \textbf{4.45\%} & \textbf{1.56} & \textbf{3.55} & \textbf{3.50\%} \\ \hline
        \hline
        \multirow{10}{*}{\rotatebox{90}{PEMS04}} & HI & 42.33 & 61.64 & 29.90\% & 42.35 & 61.66 & 29.92\% & 42.38 & 61.67 & 29.96\% & 42.35 & 61.66 & 29.92\% \\ \cline{2-14} 
        & GraphWaveNet & 17.68 & 28.53 & 12.46\% & 18.58 & 29.97 & 13.20\% & 20.01 & 31.98 & 14.22\% & 18.57 & 29.92 & 13.09\% \\ \cline{2-14}
        & DCRNN & 18.49 & 29.55 & 12.62\% & 19.58 & 31.27 & 13.32\% & 21.40 & 33.84 & 14.75\% & 19.59 & 31.28 & 13.37\% \\ \cline{2-14}
        & AGCRN & 18.41 & 29.50 & 12.49\% & 19.31 & 30.99 & 13.50\% & 20.48 & 32.75 & 14.19\% & 19.24 & 30.88 & 13.30\% \\ \cline{2-14}
        & STGCN & 18.74 & 29.94 & 13.02\% & 19.65 & 31.47 & 13.61\% & 21.27 & 33.79 & 14.73\% & 19.69 & 31.47 & 13.67\% \\ \cline{2-14}
        & StemGNN & 19.70 & 31.30 & 13.56\% & 21.83 & 34.50 & 15.15\% & 26.22 & 40.76 & 18.20\% & 22.17 & 35.12 & 15.41\% \\ \cline{2-14}
        & GTS & 19.29 & 30.49 & 13.30\% & 20.88 & 32.83 & 14.62\% & 23.56 & 36.38 & 16.92\% & 20.94 & 32.92 & 14.71\% \\ \cline{2-14}
        & MTGNN & 18.12 & 29.50 & 12.44\% & 18.97 & 31.09 & 12.91\% & 20.42 & 33.27 & 13.79\% & 18.97 & 31.04 & 12.93\% \\ \cline{2-14}
        & DGCRN & 18.85 & 29.95 & 12.92\% & 20.04 & 32.07 & 13.50\% & 22.32 & 36.28 & 14.61\% & 20.29 & 32.55 & 13.60\% \\ \cline{2-14}
        & DLinear & 22.28 & 34.95 & 14.84\% & 26.97 & 41.79 & 18.42\% & 37.37 & 56.55 & 25.47\% & 27.94 & 43.89 & 19.07\% \\ \cline{2-14}
        & PatchTST & 22.28 & 34.74 & 14.62\% & 27.06 & 41.73 & 17.79\% & 37.73 & 56.80 & 25.21\% & 28.08 & 43.89 & 18.58\% \\ \cline{2-14}
        & STID & 17.52 & 28.57 & 11.96\% & 18.34 & 29.98 & 12.48\% & 19.66 & 31.98 & 13.40\% & 18.34 & 29.95 & 12.47\% \\ \cline{2-14}
        & STNorm & 18.47 & 31.17 & 12.90\% & 19.27 & 32.99 & 13.05\% & 20.47 & 34.43 & 13.71\% & 19.21 & 32.56 & 13.10\% \\ \cline{2-14}
        & ST-MLP & \textbf{17.32} & \textbf{28.41} & \textbf{11.92\%} & \textbf{18.08} & \textbf{29.83} & \textbf{12.32\%} & \textbf{19.18} & \textbf{31.52} & \textbf{12.89\%} & \textbf{18.05} & \textbf{29.72} & \textbf{12.28\%} \\ \hline
        \hline

        \multirow{10}{*}{\rotatebox{90}{PEMS07}} & HI & 49.02 & 71.15 & 22.73\% & 49.04 & 71.18 & 22.75\% & 49.06 & 71.21 & 22.79\% & 49.03 & 71.18 & 22.75\% \\ \cline{2-14} 
        & GraphWaveNet & 18.72 & 30.62 & 7.86\% & 20.19 & 33.17 & 8.53\% & 22.46 & 36.59 & 9.65\% & 20.14 & 33.06 & 8.54\% \\ \cline{2-14}
        & DCRNN & 19.44 & 31.22 & 8.22\% & 21.12 & 34.13 & 8.93\% & 24.04 & 38.44 & 10.34\% & 21.14 & 34.13 & 8.99\% \\ \cline{2-14}
        & AGCRN & 19.35 & 31.87 & 8.17\% & 20.74 & 34.74 & 8.78\% & 22.84 & 38.32 & 9.79\% & 20.71 & 34.65 & 8.86\% \\ \cline{2-14}
        & STGCN & 20.59 & 33.30 & 8.78\% & 21.98 & 36.09 & 9.28\% & 24.51 & 40.26 & 10.72\% & 22.03 & 36.14 & 9.42\% \\ \cline{2-14}
        & StemGNN & 19.79 & 32.12 & 8.52\% & 21.91 & 35.63 & 9.42\% & 25.65 & 41.05 & 11.15\% & 22.03 & 35.81 & 9.54\% \\ \cline{2-14}
        & GTS & 20.11 & 31.92 & 8.53\% & 22.27 & 35.11 & 9.46\% & 25.72 & 39.96 & 11.08\% & 22.24 & 35.17 & 9.47\% \\ \cline{2-14}
        & MTGNN & 19.33 & 31.20 & 8.56\% & 20.93 & 34.16 & 8.90\% & 23.52 & 38.21 & 10.14\% & 20.93 & 34.14 & 9.08\% \\ \cline{2-14}
        & DGCRN & 19.03 & 30.74 & 8.16\% & 20.41 & 33.27 & 8.69\% & 22.58 & 36.74 & 9.63\% & 20.44 & 33.25 & 8.73\% \\ \cline{2-14}
        & DLinear & 24.71 & 38.05 & 9.84\% & 30.56 & 46.90 & 13.87\% & 43.70 & 65.37 & 20.42\% & 31.72 & 49.43 & 14.52\% \\ \cline{2-14}
        & PatchTST & 24.73 & 37.97 & 10.32\% & 30.75 & 47.06 & 12.92\% & 44.21 & 66.02 & 18.97\% & 31.98 & 49.68 & 13.54\% \\ \cline{2-14}
        & STID & 18.37 & 30.40 & 7.77\% & 19.64 & 32.86 & 8.29\% & 21.53 & 36.11 & 9.26\% & 19.59 & 32.77 & 8.30\% \\ \cline{2-14}
        & STNorm & 19.15 & 31.68 & 8.04\% & 20.58 & 34.88 & 8.63\% & 22.66 & 38.51 & 9.75\% & 20.50 & 34.66 & 8.69\% \\ \cline{2-14}
        & ST-MLP & \textbf{18.30} & \textbf{30.33} & \textbf{7.68\%} & \textbf{19.55} & \textbf{32.69} & \textbf{8.27\%} & \textbf{21.43} & \textbf{35.86} & \textbf{9.19\%} & \textbf{19.51} & \textbf{32.61} & \textbf{8.26\%} \\ \hline
        \hline

        \multirow{10}{*}{\rotatebox{90}{PEMS08}} & HI & 36.65 & 50.44 & 21.60\% & 36.66 & 50.45 & 21.63\% & 36.68 & 50.46 & 21.68\% & 36.66 & 50.45 & 21.63\% \\ \cline{2-14}
        & GraphWaveNet & 13.57 & 21.66 & 9.00\% & 14.48 & 23.46 & 9.33\% & 15.90 & 25.89 & 10.31\% & 14.47 & 23.46 & 9.34\% \\ \cline{2-14}
        & DCRNN & 14.07 & 22.14 & 9.23\% & 15.14 & 24.18 & 9.91\% & 16.84 & 26.87 & 11.12\% & 15.13 & 24.13 & 9.93\% \\ \cline{2-14}
        & AGCRN & 14.28 & 22.51 & 9.37\% & 15.32 & 24.41 & 9.93\% & 17.01 & 27.03 & 11.07\% & 15.35 & 24.45 & 10.01\% \\ \cline{2-14}
        & STGCN & 14.93 & 23.41 & 9.78\% & 15.92 & 25.25 & 10.19\% & 17.70 & 27.86 & 11.56\% & 15.98 & 25.26 & 10.43\% \\ \cline{2-14}
        & StemGNN & 14.98 & 23.56 & 10.64\% & 16.49 & 26.29 & 11.38\% & 19.26 & 30.37 & 13.20\% & 16.62 & 26.42 & 11.65\% \\ \cline{2-14}
        & GTS & 15.03 & 23.54 & 9.49\% & 16.48 & 26.00 & 10.50\% & 19.00 & 29.57 & 12.44\% & 16.54 & 26.02 & 10.60\% \\ \cline{2-14}
        & MTGNN & 14.23 & 22.32 & 9.98\% & 15.24 & 24.33 & 10.22\% & 16.86 & 26.85 & 12.13\% & 15.27 & 24.29 & 10.59\% \\ \cline{2-14}
        & DGCRN & 13.79 & 21.91 & 9.13\% & 14.81 & 23.83 & 9.74\% & 16.39 & 26.34 & 11.02\% & 14.85 & 23.84 & 9.84\% \\ \cline{2-14}
        & DLinear & 17.75 & 24.49 & 11.16\% & 21.62 & 34.85 & 13.33\% & 30.42 & 46.33 & 20.12\% & 22.42 & 35.44 & 14.28\% \\ \cline{2-14}
        & PatchTST & 17.81 & 27.58 & 10.76\% & 21.77 & 34.03 & 13.10\% & 30.76 & 46.86 & 18.43\% & 22.60 & 35.68 & 13.62\% \\ \cline{2-14}
        & STID & 13.32 & 21.61 & 8.79\% & 14.21 & 23.52 & 9.33\% & 15.61 & 25.91 & 10.39\% & 14.23 & 23.48 & 9.40\% \\ \cline{2-14}
        & STNorm & 14.42 & 22.85 & 9.17\% & 15.41 & 24.97 & 9.85\% & 16.88 & 27.68 & 10.70\% & 15.37 & 24.91 & 9.82\% \\ \cline{2-14}
        & ST-MLP & \textbf{13.12} & \textbf{21.29} & \textbf{8.59\%} & \textbf{14.03} & \textbf{23.10} & \textbf{9.30\%} & \textbf{15.41} & \textbf{25.48} & \textbf{10.17\%} & \textbf{14.03} & \textbf{23.07} & \textbf{9.25\%} \\ 
        \hline
    \end{tabular}}
    \end{adjustbox}
    
\caption{Results on various traffic datasets. MAE, MSE and MAPE for 3, 6, 12 horizons as well averaged over a one-hour (12 steps) are reported. Best results are bold marked. ST-MLP achieves the best performance overall.}
\label{result}
\end{table*}

\subsection{Experimental Result}

The outcomes of the model comparison are detailed in Table~\ref{result}. All baseline models exhibit significant improvements over the HI model, indicating their proficiency in capturing the underlying spatio-temporal patterns in the provided data. As the prediction horizon lengthens, the precision of all models declines, underscoring the challenge of time series forecasting as the prediction timeframe extends further into the future. 

In the case of the traffic speed dataset PEMS-BAY, our model attains the highest overall accuracy compared to other advanced STGNNs. Moreover, some basic baseline models, like STID, also demonstrate competitive results. This suggests that crafting effective traffic forecasting algorithms might not always need intricate neural network architectures.

Concerning the traffic flow datasets PEMS04, PEMS07, and PEMS08, our model exhibits superior performance and consistently surpasses the baseline methods by a considerable margin. It's worth noting that approaches specifically tailored for multivariate time series forecasting (DLinear, PatchTST), without incorporating spatial characteristics, understandably exhibit diminished performance in the context of spatio-temporal forecasting.

\subsection{Efficiency Study compared to STGNNs methods}

Since ST-MLP outperforms all the STGNNs method, in this section, we conduct a comparison of the efficiency of ST-MLP with other STGNNs methods using all datasets. To ensure a more intuitive and effective comparison, we analyze the average training time required for one epoch of these models. To get a better visualization in various datasets, in each dataset we normalize the highest time consumption to 1, then the time of other models are also normalized correspondingly.

\begin{figure}[t]
\begin{center}
\includegraphics[width = \linewidth]{ 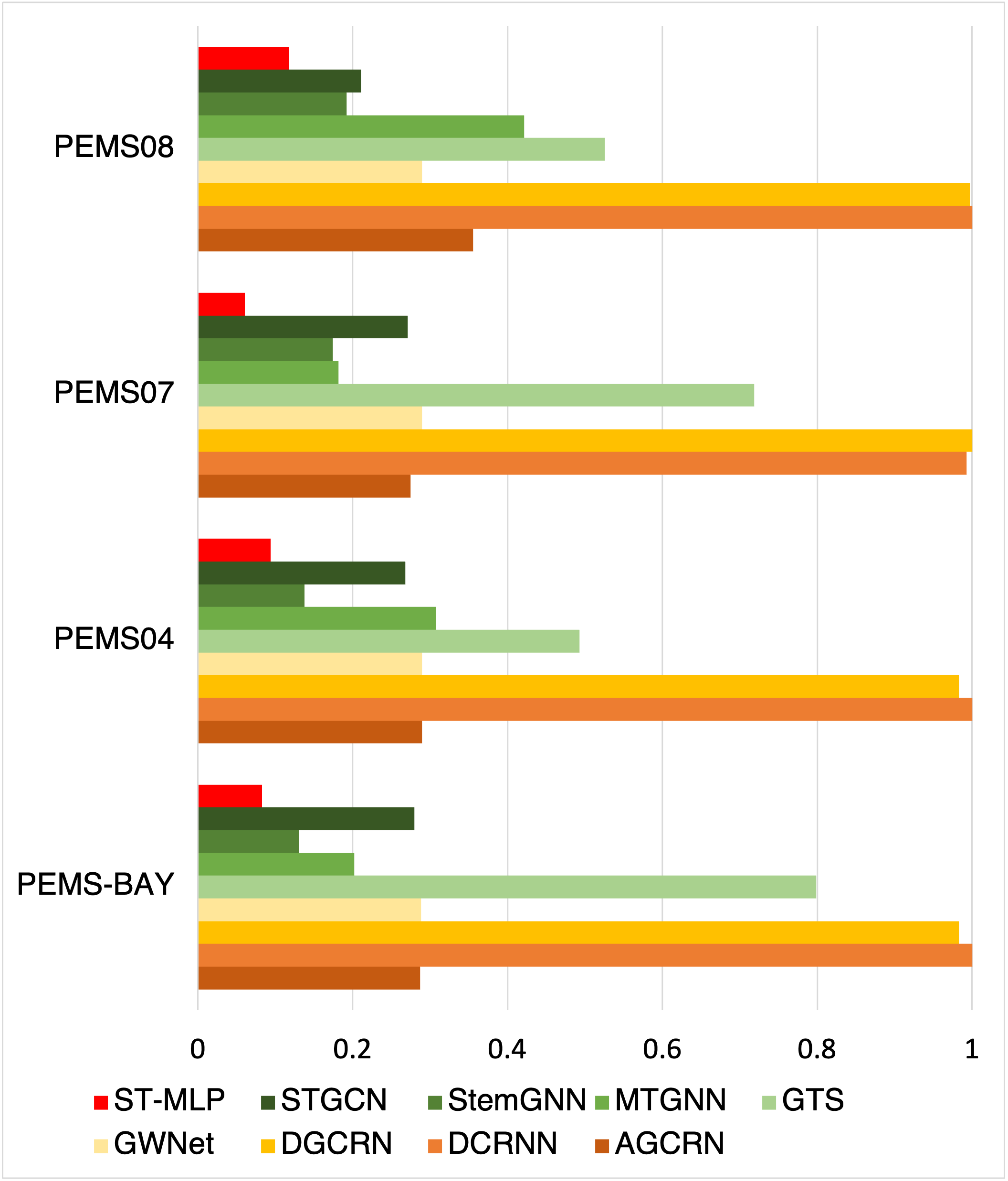}
\end{center}
\caption{Normalized training time for one epoch. The slowest model is normalized to 1 on each experiment dataset. ST-MLP requires significantly less training time than other STGNNs counterparts on all datasets.}
\label{fig:efficiency}
\end{figure}

The normalized training time consumption is illustrated in Figure~\ref{fig:efficiency}. In contrast to other STGNN methods, our ST-MLP achieves exceptional computational speed benefiting from its simplicity. This underscores the efficiency of our proposed approach. Among the STGNN methods, StemGNN and MTGNN demonstrate relatively favorable computational efficiency, although their accuracy is not exceptional.

\subsection{Ablation Study}

Due to space constraints, the ablation study primarily presents experiment results conducted on specific datasets. Nevertheless, it's important to acknowledge that similar conclusions and analyses are applicable to PEMS-BAY, PEMS04, PEMS07, and PEMS08 datasets as well.

\subsubsection{Importance of Different Embedding Components}

\begin{figure}[t]
\begin{center}
\includegraphics[width = 0.9\linewidth]{ 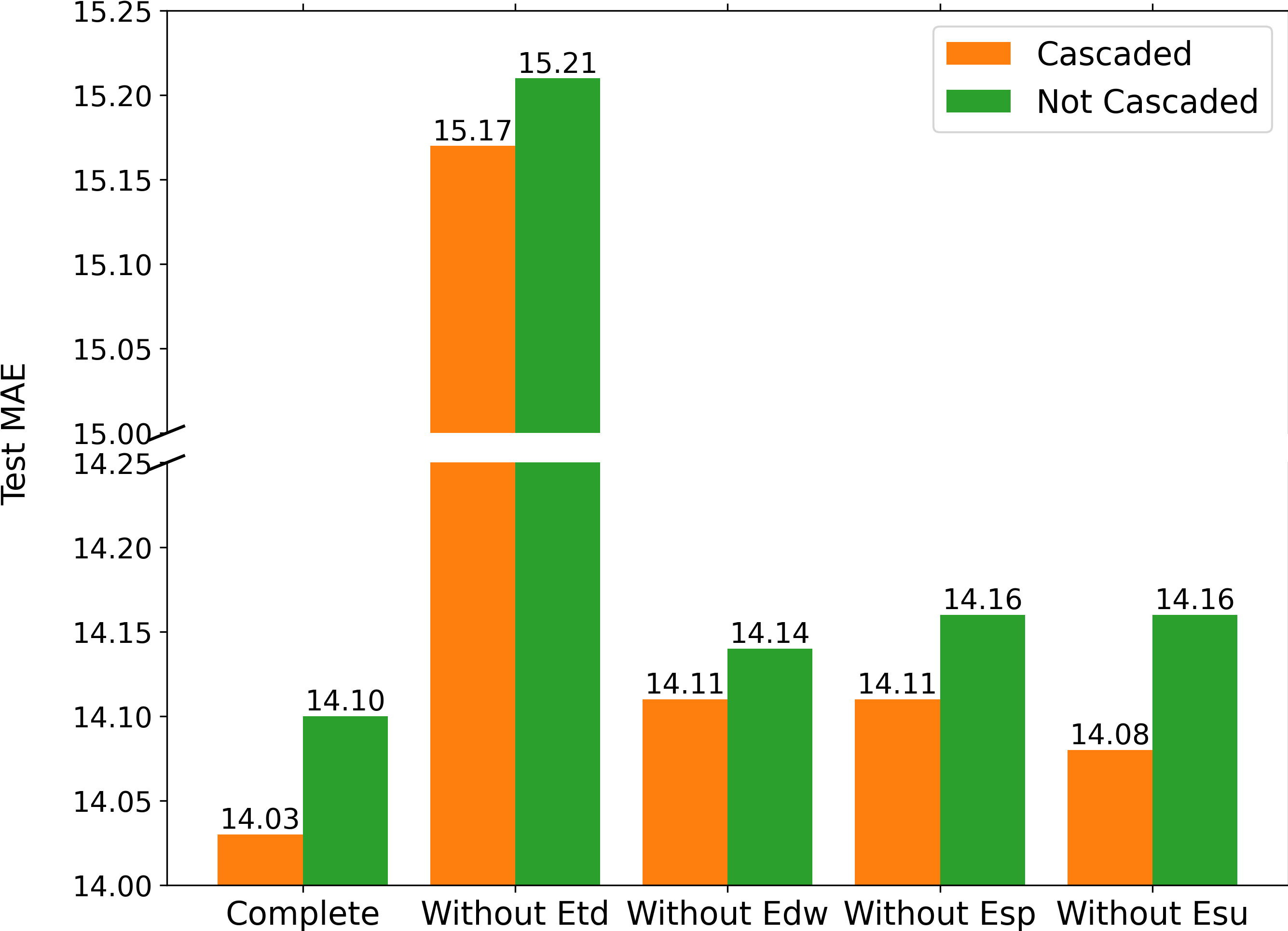}
\end{center}
\caption{Test MAE of variants of ST-MLP on PEMS08. The cascaded model with complete embeddings achieves the lowest test error among all variants.}
\label{fig: ablation 1}
\end{figure}

In this section, we conduct ablation studies to assess the effectiveness of the four different embeddings in our ST-MLP model. Specifically, we examine four variants of the ST-MLP: one without Time in Day information ($E_{td}$), one without Day in Week information ($E_{dw}$), one without predefined graph information ($E_{sp}$), and one without the unknown spatial embedding ($E_{su}$). For each variant, we analyze its performance in comparison to the complete ST-MLP model. By systematically removing specific embeddings, our goal is to discern their individual contributions to the overall forecasting accuracy. The test MAE values of these ST-MLP variants are illustrated in Figure~\ref{fig: ablation 1}.

In summary, the ablation studies demonstrate that all embeddings play a positive role in improving the final forecasting accuracy. Among the four embeddings analyzed, Time in Day ($E_{td}$) emerges as the most crucial feature, indicating that capturing the daily seasonal traffic trends is essential for achieving accurate forecasts. 

In this framework, the combination of predefined graph connections and the exploration of more spatial features contributes to improved accuracy. Leveraging spatial information through graph connections and additional features enhances the model's capability to capture traffic patterns, thereby further enhancing the accuracy of traffic forecasting.

\subsubsection{Importance of Cascaded Structure}

To demonstrate the efficacy of the cascaded structure, we propose other variants of ST-MLP which are not cascaded, denoted by the green bars in Figure~\ref{fig: ablation 1}. 
Rather than adopting sequential aggregation of \(E_t\), \(E_s\), and \(E_d\) as depicted in Figure~\ref{fig:model} (a), we opt for a parallel concatenation strategy: \(E = \texttt{Concat}[(E_t, E_s, E_d), \texttt{dim}=1]\). Subsequently, we employ a linear layer to transform this into the forecast result \(\hat{X} \in R^{N \times Q}\).

Our results underscore the important role of the cascaded structure in bolstering the model's performance. Remarkably, the complete cascaded model demonstrates the best performance, while a consistent improvement is observed in the cascaded model compared to its non-cascaded counterparts, even when specific embeddings are absent. This lends empirical support to the effectiveness of our cascaded design across a broad range of cases.

\subsubsection{Ablation on Channel-independence}



Typically, spatial information is commonly utilized by mixing channels, such as Graph Neural Networks (GNNs) or Graph Convolutional Networks (GCNs), which is opposite to CI. To investigate the impact of CI vs CM strategy, we introduce a variant of ST-MLP. We add a simple linear layer before the forecast in Figure~\ref{fig:model} (a) to mix the channels, which belongs to the \texttt{ChannelMix} module type in Equation (\ref{cmixing}), creating a new variant named "CM ST-MLP". We then recorded the metrics on the training, validation, and testing datasets of PEMS-BAY, PEMS04, and PEMS08. The results are presented in Table~\ref{ab2}.

The outcomes demonstrate a potential issue with the CM ST-MLP model. Despite having similar validation errors compared to the original CI ST-MLP model, CM ST-MLP exhibits lower training error but much higher testing error. This more severe overfitting problem indicates that even a simple linear operation to mix channels can negatively impact the model's generalization ability when compared to the CI training strategy used in ST-MLP.

The overfitting issue observed in the CM ST-MLP model can be attributed to a couple of potential reasons, primarily the distribution shift problem, where the difference between the train and test distributions could degrade model performance
As we observed in Figure \ref{fig:distribution}, in real-world applications, the existence of non-stationary time series is quite common, which makes robustness against distribution shift crucial. Recent studies reveal that the choice between CM and CI is a trade-off between robustness and capacity~\cite{han2023ci}. While CM models are more expressive, they might inadvertently capture false spatio-temporal correlations due to the mixing on spatial channel. On the other hand, the CI training strategy in our ST-MLP could alleviate the problem of data distribution shift, leading to improved generalization and more accurate traffic forecasting.



\section{Conclusion and Future Work}
\label{Conclusion}

In this study, we have introduced a novel model for accurate traffic forecasting, termed ST-MLP. By leveraging the CI strategy and simple MLPs, our framework achieves competitive forecasting accuracy. The evaluation metrics we employ underscore the effectiveness of our approach. We anticipate that this work will inspire the exploration of straightforward models in the domains of traffic and spatio-temporal forecasting, offering valuable insights for both neural architecture understanding and practical applications.

To expand upon our research, it would be worthwhile to extend our analysis to diverse spatio-temporal datasets and delve into enhancing robustness using CI strategies. Additionally, considering the faster convergence rate and mitigation of over-smoothing associated with CI~\cite{nie2022time, Zhou2022K} present interesting avenues for future exploration.

\begin{table}[t]
\centering
\renewcommand{\arraystretch}{1.2}
\begin{adjustbox}{width=0.47\textwidth}
{\fontsize{10}{12}\selectfont
\begin{tabular}{|c|c|c|c|c|c|c|c|}
\hline
\multicolumn{1}{|c|}{\multirow{2}{*}{\centering Dataset}} & \multicolumn{1}{|c|}{\multirow{2}{*}{\centering Stage}} & \multicolumn{3}{c|}{CI ST-MLP} & \multicolumn{3}{c|}{CM ST-MLP} \\
\cline{3-8}
 &  & MAE & RMSE & MAPE & MAE & RMSE & MAPE \\
\hline
 \multirow{3}{*}{\centering PEMS-BAY} & Training & 1.41  & 3.11  & 3.05\%  & 1.40  & 3.04  & 3.06\%  \\
\cline{2-8}
 & Validation & 1.56  & 3.31  & 3.58\%  & 1.68  & 3.39  & 3.86\%  \\
\cline{2-8}
 & Testing & 1.56  & 3.55  & 3.50\%  & 1.75  & 3.73  & 3.92\%  \\
\hline
 \multirow{3}{*}{\centering PEMS04} & Training & 16.75 & 27.61 & 12.60\% & 16.52 & 27.15 & 12.95\% \\
\cline{2-8}
 & Validation & 17.76 & 28.40 & 12.18\% & 19.41 & 30.44 & 13.90\% \\
\cline{2-8}
 & Testing & 18.05 & 29.72 & 12.28\% & 20.74 & 33.85 & 14.60\% \\
\hline
 \multirow{3}{*}{\centering PEMS08} & Training & 14.11 & 23.54 & 9.64\% & 13.25 & 21.34 & 8.99\% \\
\cline{2-8}
 & Validation & 14.11 & 23.19 & 10.67\% & 14.12 & 22.45 & 9.78\% \\
\cline{2-8}
 & Testing & 14.03 & 23.08 & 9.22\% & 15.58 & 25.21 & 10.34\% \\
\hline
\end{tabular}}
\end{adjustbox}
\caption{CI vs CM strategies on ST-MLP. The CM ST-MLP model displays lower training error but higher testing error compared to the CI variant}
\label{ab2}
\end{table}


\section{Related Works}

\label{Literature Review}

\subsection{Traffic Forecasting}

As a key component of ITS, traffic forecasting has been the subject of study for decades. As early as 1979, AutoRegressive Integrated Moving Average (ARIMA) models are utilized to forecast freeway traffic data~\cite{williams2003modeling}. Over time, numerous machine learning-based methods have been explored for this task, including Support Vector Machine (SVM)~\cite{zhang2009traffic}, Decision Tree~\cite{xia2017traffic}, Bayesian Networks~\cite{sun2006bayesian}, among others. 

Deep learning has revolutionized traffic forecasting algorithms, offering greater flexibility and complexity by designing neural structures for specific functions in different layers. Long Short-Term Memory (LSTM) neural networks have been effectively applied to tackle traffic forecasting tasks, leveraging their ability to handle sequential data~\cite{zhao2017lstm}. Another approach by \textit{Zhang et al.} involved representing the traffic network as a regular 2D grid and using traditional Convolutional Neural Networks (CNNs) to forecast traffic data~\cite{zhang2019short}.

Graph neural networks (GNNs) are well-suited for traffic research, as road networks can be effectively represented by graph structures \cite{Chen2021ZGCNETsTZ}. For instance, some papers utilize the attention mechanism to analyze spatial and temporal correlations in traffic data~\cite{guo2019attention,zheng2020gman}. \textit{Xu et al.} leverage the combination of graph structure and transformer framework to achieve high forecast accuracy~\cite{xu2020spatial}. Moreover, several models have explored dynamic graph structures to better capture temporal-spatial relationships, moving beyond fixed and predefined road graphs~\cite{zhang2020spatio,guo2021learning}. These approaches demonstrate the effectiveness of GNNs in incorporating spatio-temporal information for improved traffic forecasting performance.


\subsection{Linear Models for Time Series Forecasting}

Recent years have seen a surge in research in the field of time series forecast, particularly the transformer-based models, including Informer~\cite{zhou2021informer}, Autoformer~\cite{wu2021autoformer}, Fedformer~\cite{zhou2022fedformer}, Pyraformer~\cite{liu2021pyraformer}, etc. However, \textit{Zeng et al.} demonstrated that using some simple linear models can outperform many transformer architectures in long-term time series forecasting tasks~\cite{zeng2023transformers}. Since MLP structures have less complexity and computational cost, they have also garnered attention from researchers. For instance, \textit{Ekambaram et al.} followed the idea of MLP mixer in computer vision~\cite{tolstikhin2021mlp} and performed mixing operations along both the time and feature dimensions in their TSMixer model~\cite{vijay2023tsmixer}. There are also some more emerging MLP-based designs like Koopa~\cite{liu2023KoopaLN} and FITS~\cite{xu2023FITS}, further highlighting the potential of MLP-based approaches in time series forecasting.




\bibliography{aaai24}

\begin{thebibliography}{52}
\providecommand{\natexlab}[1]{#1}

\bibitem[{Bai et~al.(2020)Bai, Yao, Li, Wang, and Wang}]{bai2020adaptive}
Bai, L.; Yao, L.; Li, C.; Wang, X.; and Wang, C. 2020.
\newblock Adaptive Graph Convolutional Recurrent Network for Traffic
  Forecasting.
\newblock \emph{Advances in Neural Information Processing Systems}, 33:
  17804--17815.

\bibitem[{Cao et~al.(2020)Cao, Wang, Duan, Zhang, Zhu, Huang, Tong, Xu, Bai,
  Tong et~al.}]{cao2020spectral}
Cao, D.; Wang, Y.; Duan, J.; Zhang, C.; Zhu, X.; Huang, C.; Tong, Y.; Xu, B.;
  Bai, J.; Tong, J.; et~al. 2020.
\newblock Spectral Temporal Graph Neural Network for Multivariate Time-series
  Forecasting.
\newblock \emph{Advances in Neural Information Processing Systems}, 33:
  17766--17778.

\bibitem[{Chen et~al.(2001)Chen, Petty, Skabardonis, Varaiya, and
  Jia}]{chen2001pems}
Chen, C.; Petty, K.~F.; Skabardonis, A.; Varaiya, P.~P.; and Jia, Z. 2001.
\newblock Freeway Performance Measurement System: Mining Loop Detector Data.
\newblock \emph{Transportation Research Record}, 1748: 102 -- 96.

\bibitem[{Chen, Segovia-Dominguez, and Gel(2021)}]{Chen2021ZGCNETsTZ}
Chen, Y.; Segovia-Dominguez, I.; and Gel, Y.~R. 2021.
\newblock Z-gcnets: Time Zigzags at Graph Convolutional Networks for Time
  Series Forecasting.
\newblock In \emph{Proceedings of the International Conference on Machine
  Learning}.

\bibitem[{Cui, Xie, and Zheng(2021)}]{cui2021historical}
Cui, Y.; Xie, J.; and Zheng, K. 2021.
\newblock Historical Inertia: A Neglected But Powerful Baseline for Long
  Sequence Time-series Forecasting.
\newblock In \emph{Proceedings of the 30th ACM International Conference on
  Information \& Knowledge Management}, 2965--2969.

\bibitem[{Das et~al.(2023)Das, Kong, Leach, Sen, and Yu}]{das2023long}
Das, A.; Kong, W.; Leach, A.; Sen, R.; and Yu, R. 2023.
\newblock Long-term Forecasting with TiDE: Time-series Dense Encoder.
\newblock arXiv:2304.08424.

\bibitem[{Deng et~al.(2021)Deng, Chen, Jiang, Song, and
  Tsang}]{providence2022spatial}
Deng, J.; Chen, X.; Jiang, R.; Song, X.; and Tsang, I.~W. 2021.
\newblock St-norm: Spatial and temporal normalization for multi-variate time
  series forecasting.
\newblock In \emph{Proceedings of the 27th ACM SIGKDD Conference on Knowledge
  Discovery \& Data Mining}, 269--278.

\bibitem[{Ekambaram et~al.(2023)Ekambaram, Jati, Nguyen, Sinthong, and
  Kalagnanam}]{vijay2023tsmixer}
Ekambaram, V.; Jati, A.; Nguyen, N.; Sinthong, P.; and Kalagnanam, J. 2023.
\newblock TSMixer: Lightweight MLP-mixer Model for Multivariate Time Series
  Forecasting.
\newblock arXiv:2306.09364.

\bibitem[{Guo et~al.(2019)Guo, Lin, Feng, Song, and Wan}]{guo2019attention}
Guo, S.; Lin, Y.; Feng, N.; Song, C.; and Wan, H. 2019.
\newblock Attention Based Spatial-temporal Graph Convolutional Networks for
  Traffic Flow Forecasting.
\newblock In \emph{Proceedings of the AAAI Conference on Artificial
  Intelligence}, 01, 922--929.

\bibitem[{Guo et~al.(2021)Guo, Lin, Wan, Li, and Cong}]{guo2021learning}
Guo, S.; Lin, Y.; Wan, H.; Li, X.; and Cong, G. 2021.
\newblock Learning Dynamics And Heterogeneity of Spatial-temporal Graph Data
  for Traffic Forecasting.
\newblock \emph{IEEE Transactions on Knowledge and Data Engineering}, 34(11):
  5415--5428.

\bibitem[{Han, Ye, and Zhan(2023)}]{han2023ci}
Han, L.; Ye, H.-J.; and Zhan, D.-C. 2023.
\newblock The Capacity And Robustness Trade-off: Revisiting the Channel
  Independent Strategy for Multivariate Time Series Forecasting.
\newblock arXiv:2304.05206.

\bibitem[{Jin et~al.(2023)Jin, Koh, Wen, Zambon, Alippi, Webb, King, and
  Pan}]{jin2023survey}
Jin, M.; Koh, H.~Y.; Wen, Q.; Zambon, D.; Alippi, C.; Webb, G.~I.; King, I.;
  and Pan, S. 2023.
\newblock A Survey on Graph Neural Networks for Time Series: Forecasting,
  Classification, Imputation, and Anomaly Detection.
\newblock arXiv:2307.03759.

\bibitem[{Kim et~al.(2021)Kim, Kim, Tae, Park, Choi, and Choo}]{kim2021revin}
Kim, T.; Kim, J.; Tae, Y.; Park, C.; Choi, J.-H.; and Choo, J. 2021.
\newblock Reversible Instance Normalization for Accurate Time-Series
  Forecasting against Distribution Shift.
\newblock In \emph{Proceedings of the International Conference on Learning
  Representations}.

\bibitem[{Kipf and Welling(2016)}]{kipf2016semi}
Kipf, T.~N.; and Welling, M. 2016.
\newblock Semi-supervised Classification with Graph Convolutional Networks.
\newblock arXiv:1609.02907.

\bibitem[{Li et~al.(2023)Li, Feng, Yan, Jin, Yang, Sun, Jin, and
  Li}]{li2023dynamic}
Li, F.; Feng, J.; Yan, H.; Jin, G.; Yang, F.; Sun, F.; Jin, D.; and Li, Y.
  2023.
\newblock Dynamic Graph Convolutional Recurrent Network for Traffic Prediction:
  Benchmark And Solution.
\newblock \emph{ACM Transactions on Knowledge Discovery from Data}, 17(1):
  1--21.

\bibitem[{Li et~al.(2017)Li, Yu, Shahabi, and Liu}]{li2017diffusion}
Li, Y.; Yu, R.; Shahabi, C.; and Liu, Y. 2017.
\newblock Diffusion Convolutional Recurrent Neural Network: Data-driven Traffic
  Forecasting.
\newblock arXiv:1707.01926.

\bibitem[{Liang et~al.(2023)Liang, Shao, Wang, Zhang, Sun, and
  Xu}]{shao2022basicts}
Liang, Y.; Shao, Z.; Wang, F.; Zhang, Z.; Sun, T.; and Xu, Y. 2023.
\newblock Basicts: An Open Source Fair Multivariate Time Series Prediction
  Benchmark.
\newblock In \emph{Benchmarking, Measuring, and Optimizing}, 87--101. Springer
  International Publishing.

\bibitem[{Liu et~al.(2021)Liu, Yu, Liao, Li, Lin, Liu, and
  Dustdar}]{liu2021pyraformer}
Liu, S.; Yu, H.; Liao, C.; Li, J.; Lin, W.; Liu, A.~X.; and Dustdar, S. 2021.
\newblock Pyraformer: Low-complexity Pyramidal Attention for Long-range Time
  Series Modeling And Forecasting.
\newblock In \emph{Proceedings of the International Conference on Learning
  Representations}.

\bibitem[{Liu et~al.(2023{\natexlab{a}})Liu, Liang, Huang, Hu, Cao, Hooi, and
  Zimmermann}]{liu2023we}
Liu, X.; Liang, Y.; Huang, C.; Hu, H.; Cao, Y.; Hooi, B.; and Zimmermann, R.
  2023{\natexlab{a}}.
\newblock Do We Really Need Graph Neural Networks for Traffic Forecasting?
\newblock arXiv:2301.12603.

\bibitem[{Liu et~al.(2023{\natexlab{b}})Liu, Li, Wang, and
  Long}]{liu2023KoopaLN}
Liu, Y.; Li, C.; Wang, J.; and Long, M. 2023{\natexlab{b}}.
\newblock Koopa: Learning Non-stationary Time Series Dynamics with Koopman
  Predictors.
\newblock arXiv:2305.18803.

\bibitem[{Nie et~al.(2023)Nie, H.~Nguyen, Sinthong, and
  Kalagnanam}]{nie2022time}
Nie, Y.; H.~Nguyen, N.; Sinthong, P.; and Kalagnanam, J. 2023.
\newblock A Time Series is Worth 64 Words: Long-term Forecasting with
  Transformers.
\newblock In \emph{Proceedings of the International Conference on Learning
  Representations}.

\bibitem[{Oreshkin et~al.(2021)Oreshkin, Amini, Coyle, and
  Coates}]{oreshkin2021fc}
Oreshkin, B.~N.; Amini, A.; Coyle, L.; and Coates, M. 2021.
\newblock Fc-gaga: Fully Connected Gated Graph Architecture for Spatio-temporal
  Traffic Forecasting.
\newblock In \emph{Proceedings of the AAAI Conference on Artificial
  Intelligence}, volume~35, 9233--9241.

\bibitem[{Qin et~al.(2023)Qin, Luo, Zhao, Fang, Tao, and Wang}]{qin2023mlp}
Qin, Y.; Luo, H.; Zhao, F.; Fang, Y.; Tao, X.; and Wang, C. 2023.
\newblock Spatio-temporal hierarchical MLP network for traffic forecasting.
\newblock \emph{Information Sciences}, 632: 543--554.

\bibitem[{Shang, Chen, and Bi(2021)}]{shang2021discrete}
Shang, C.; Chen, J.; and Bi, J. 2021.
\newblock Discrete Graph Structure Learning for Forecasting Multiple Time
  Series.
\newblock arXiv:2101.06861.

\bibitem[{Shao et~al.(2022{\natexlab{a}})Shao, Zhang, Wang, Wei, and
  Xu}]{shao2022spatial}
Shao, Z.; Zhang, Z.; Wang, F.; Wei, W.; and Xu, Y. 2022{\natexlab{a}}.
\newblock Spatial-temporal Identity: A Simple Yet Effective Baseline for
  Multivariate Time Series Forecasting.
\newblock In \emph{Proceedings of the 31st ACM International Conference on
  Information \& Knowledge Management}, 4454--4458.

\bibitem[{Shao et~al.(2022{\natexlab{b}})Shao, Zhang, Wei, Wang, Xu, Cao, and
  Jensen}]{shao2022decoupled}
Shao, Z.; Zhang, Z.; Wei, W.; Wang, F.; Xu, Y.; Cao, X.; and Jensen, C.~S.
  2022{\natexlab{b}}.
\newblock Decoupled Dynamic Spatial-temporal Graph Neural Network for Traffic
  Forecasting.
\newblock arXiv:2206.09112.

\bibitem[{Sun, Zhang, and Yu(2006)}]{sun2006bayesian}
Sun, S.; Zhang, C.; and Yu, G. 2006.
\newblock A Bayesian Network Approach to Traffic Flow Forecasting.
\newblock \emph{IEEE Transactions on Intelligent Transportation Systems}, 7(1):
  124--132.

\bibitem[{Tolstikhin et~al.(2021)Tolstikhin, Houlsby, Kolesnikov, Beyer, Zhai,
  Unterthiner, Yung, Steiner, Keysers, Uszkoreit et~al.}]{tolstikhin2021mlp}
Tolstikhin, I.~O.; Houlsby, N.; Kolesnikov, A.; Beyer, L.; Zhai, X.;
  Unterthiner, T.; Yung, J.; Steiner, A.; Keysers, D.; Uszkoreit, J.; et~al.
  2021.
\newblock MLP-mixer: An all-MLP Architecture for Vision.
\newblock \emph{Advances in Neural Information Processing Systems}, 34:
  24261--24272.

\bibitem[{Wang(2020)}]{wang2020easytorch}
Wang, Y. 2020.
\newblock {EasyTorch}: Simple and Powerful Pytorch Framework.
\newblock \url{https://github.com/cnstark/easytorch}.

\bibitem[{Wang, Sun, and Boukerche(2022)}]{wang2022novel}
Wang, Z.; Sun, P.; and Boukerche, A. 2022.
\newblock A Novel Time Efficient Machine Learning-based Traffic Flow Prediction
  Method for Large Scale Road Network.
\newblock In \emph{Proceedings of the 2022 IEEE International Conference on
  Communications}, 3532--3537. IEEE.

\bibitem[{Wang et~al.(2022{\natexlab{a}})Wang, Sun, Hu, and
  Boukerche}]{wang2022novel1}
Wang, Z.; Sun, P.; Hu, Y.; and Boukerche, A. 2022{\natexlab{a}}.
\newblock A Novel Mixed Method of Machine Learning Based Models in Vehicular
  Traffic Flow Prediction.
\newblock In \emph{Proceedings of the 25th International ACM Conference on
  Modeling Analysis And Simulation of Wireless And Mobile Systems}, 95--101.

\bibitem[{Wang et~al.(2022{\natexlab{b}})Wang, Sun, Hu, and
  Boukerche}]{wang2022sfl}
Wang, Z.; Sun, P.; Hu, Y.; and Boukerche, A. 2022{\natexlab{b}}.
\newblock SFL: A High-precision Traffic Flow Predictor for Supporting
  Intelligent Transportation Systems.
\newblock In \emph{Proceedings of the 2022 IEEE Global Communications
  Conference}, 251--256. IEEE.

\bibitem[{Wang et~al.(2023{\natexlab{a}})Wang, Sun, Hu, and
  Boukerche}]{wang2023novel}
Wang, Z.; Sun, P.; Hu, Y.; and Boukerche, A. 2023{\natexlab{a}}.
\newblock A Novel Hybrid Method for Achieving Accurate And Timeliness Vehicular
  Traffic Flow Prediction in Road Networks.
\newblock \emph{Computer Communications}, 209: 378--386.

\bibitem[{Wang et~al.(2023{\natexlab{b}})Wang, Zhuang, Li, Zhao, and
  Sun}]{wang2023st}
Wang, Z.; Zhuang, D.; Li, Y.; Zhao, J.; and Sun, P. 2023{\natexlab{b}}.
\newblock ST-GIN: An Uncertainty Quantification Approach in Traffic Data
  Imputation with Spatio-temporal Graph Attention And Bidirectional Recurrent
  United Neural Networks.
\newblock arXiv:2305.06480.

\bibitem[{Weng et~al.(2023)Weng, Fan, Wu, Hu, Tian, Zhu, and
  Wu}]{weng2023decomposition}
Weng, W.; Fan, J.; Wu, H.; Hu, Y.; Tian, H.; Zhu, F.; and Wu, J. 2023.
\newblock A Decomposition Dynamic Graph Convolutional Recurrent Network for
  Traffic Forecasting.
\newblock \emph{Pattern Recognition}, 142: 109670.

\bibitem[{Williams and Hoel(2003)}]{williams2003modeling}
Williams, B.~M.; and Hoel, L.~A. 2003.
\newblock Modeling and Forecasting Vehicular Traffic Flow as a Seasonal ARIMA
  Process: Theoretical Basis And Empirical Results.
\newblock \emph{Journal of Transportation Engineering}, 129(6): 664--672.

\bibitem[{Wu et~al.(2021)Wu, Xu, Wang, and Long}]{wu2021autoformer}
Wu, H.; Xu, J.; Wang, J.; and Long, M. 2021.
\newblock Autoformer: Decomposition Transformers with Auto-correlation for
  Long-term Series Forecasting.
\newblock \emph{Advances in Neural Information Processing Systems}, 34:
  22419--22430.

\bibitem[{Wu et~al.(2020)Wu, Pan, Long, Jiang, Chang, and
  Zhang}]{wu2020connecting}
Wu, Z.; Pan, S.; Long, G.; Jiang, J.; Chang, X.; and Zhang, C. 2020.
\newblock Connecting the Dots: Multivariate Time Series Forecasting with Graph
  Neural Networks.
\newblock In \emph{Proceedings of the 26th ACM SIGKDD International Conference
  on Knowledge Discovery \& Data Mining}, 753--763.

\bibitem[{Wu et~al.(2019)Wu, Pan, Long, Jiang, and Zhang}]{wu2019graph}
Wu, Z.; Pan, S.; Long, G.; Jiang, J.; and Zhang, C. 2019.
\newblock Graph Wavenet for Deep Spatial-temporal Graph Modeling.
\newblock arXiv:1906.00121.

\bibitem[{Xia and Chen(2017)}]{xia2017traffic}
Xia, Y.; and Chen, J. 2017.
\newblock Traffic Flow Forecasting Method Based on Gradient Boosting Decision
  Tree.
\newblock In \emph{Proceedings of the 5th International Conference on Frontiers
  of Manufacturing Science and Measuring Technology}, 413--416. Atlantis Press.

\bibitem[{Xu et~al.(2020)Xu, Dai, Liu, Gao, Lin, Qi, and Xiong}]{xu2020spatial}
Xu, M.; Dai, W.; Liu, C.; Gao, X.; Lin, W.; Qi, G.-J.; and Xiong, H. 2020.
\newblock Spatial-temporal Transformer Networks for Traffic Flow Forecasting.
\newblock arXiv:2001.02908.

\bibitem[{Xu, Zeng, and Xu(2023)}]{xu2023FITS}
Xu, Z.; Zeng, A.; and Xu, Q. 2023.
\newblock Fits: Modeling Time Series with 10k Parameters.
\newblock arXiv:2307.03756.

\bibitem[{Yu, Yin, and Zhu(2017)}]{yu2017spatio}
Yu, B.; Yin, H.; and Zhu, Z. 2017.
\newblock Spatio-temporal Graph Convolutional Networks: A Deep Learning
  Framework for Traffic Forecasting.
\newblock arXiv:1709.04875.

\bibitem[{Zeng et~al.(2023)Zeng, Chen, Zhang, and Xu}]{zeng2023transformers}
Zeng, A.; Chen, M.; Zhang, L.; and Xu, Q. 2023.
\newblock Are Transformers Effective for Time Series Forecasting?
\newblock In \emph{Proceedings of the AAAI Conference on Artificial
  Intelligence}, 9, 11121--11128.

\bibitem[{Zhang et~al.(2020)Zhang, Chang, Meng, Xiang, and
  Pan}]{zhang2020spatio}
Zhang, Q.; Chang, J.; Meng, G.; Xiang, S.; and Pan, C. 2020.
\newblock Spatio-temporal Graph Structure Learning for Traffic Forecasting.
\newblock In \emph{Proceedings of the AAAI Conference on Artificial
  Intelligence}, 01, 1177--1185.

\bibitem[{Zhang et~al.(2019)Zhang, Yu, Qi, Shu, and Wang}]{zhang2019short}
Zhang, W.; Yu, Y.; Qi, Y.; Shu, F.; and Wang, Y. 2019.
\newblock Short-term Traffic Flow Prediction Based on Spatio-temporal Analysis
  And Cnn Deep Learning.
\newblock \emph{Transportmetrica A: Transport Science}, 15(2): 1688--1711.

\bibitem[{Zhang and Liu(2009)}]{zhang2009traffic}
Zhang, Y.; and Liu, Y. 2009.
\newblock Traffic Forecasting Using Least Squares Support Vector Machines.
\newblock \emph{Transportmetrica}, 5(3): 193--213.

\bibitem[{Zhao et~al.(2017)Zhao, Chen, Wu, Chen, and Liu}]{zhao2017lstm}
Zhao, Z.; Chen, W.; Wu, X.; Chen, P.~C.; and Liu, J. 2017.
\newblock Lstm Network: A Deep Learning Approach for Short-term Traffic
  Forecast.
\newblock \emph{IET Intelligent Transport Systems}, 11(2): 68--75.

\bibitem[{Zheng et~al.(2020)Zheng, Fan, Wang, and Qi}]{zheng2020gman}
Zheng, C.; Fan, X.; Wang, C.; and Qi, J. 2020.
\newblock GMAN: A Graph Multi-attention Network for Traffic Prediction.
\newblock In \emph{Proceedings of the AAAI Conference on Artificial
  Intelligence}, volume~34, 1234--1241.

\bibitem[{Zhou et~al.(2021)Zhou, Zhang, Peng, Zhang, Li, Xiong, and
  Zhang}]{zhou2021informer}
Zhou, H.; Zhang, S.; Peng, J.; Zhang, S.; Li, J.; Xiong, H.; and Zhang, W.
  2021.
\newblock Informer: Beyond Efficient Transformer for Long Sequence Time-series
  Forecasting.
\newblock In \emph{Proceedings of the AAAI Conference on Artificial
  Intelligence}, 12, 11106--11115.

\bibitem[{Zhou et~al.(2022{\natexlab{a}})Zhou, Ma, Wen, Wang, Sun, and
  Jin}]{zhou2022fedformer}
Zhou, T.; Ma, Z.; Wen, Q.; Wang, X.; Sun, L.; and Jin, R. 2022{\natexlab{a}}.
\newblock Fedformer: Frequency Enhanced Decomposed Transformer for Long-term
  Series Forecasting.
\newblock In \emph{Proceedings of the International Conference on Machine
  Learning}, 27268--27286. PMLR.

\bibitem[{Zhou et~al.(2022{\natexlab{b}})Zhou, Zhong, Yang, Wang, Yang, and
  Shen}]{Zhou2022K}
Zhou, Z.; Zhong, R.; Yang, C.; Wang, Y.; Yang, X.; and Shen, W.
  2022{\natexlab{b}}.
\newblock A K-variate Time Series Is Worth K Words: Evolution of the Vanilla
  Transformer Architecture for Long-term Multivariate Time Series Forecasting.
\newblock arXiv:2212.02789.

\end{thebibliography}

\newpage
~
\newpage

\appendix

\section{Appendix}

\subsection{Notation}

In this section, we provide a comprehensive definition about all notation in the section \textbf{Proposed Method}, summarized in Table~\ref{Notations}.

\begin{table}[!ht]
\begin{center}
\renewcommand{\arraystretch}{0.9}
\setlength{\tabcolsep}{8pt}
\begin{tabular}{|l|p{0.7\linewidth}|}
\hline
Notation & Description \\
\hline
$A$ & Predefined graph matrix \\
\hline
$B_{td}$ & Codebook of Time in Day (TD) \\
\hline
$B_{dw}$ & Codebook of Day in Week (DW) \\
\hline
$B_{sp}$ & Codebook of predefined graph structure \\
\hline
$B_{su}$ & Codebook of unknown spatial information \\
\hline
$E_t$ & Temporal embedding \\
\hline
$E_{td}$ & Temporal embedding of TD \\
\hline
$E_{td}^i$ & $E_{td}$ for node $i$  \\
\hline
$E_{dw}$ & Temporal embedding of DW \\
\hline
$E_{dw}^i$ & $E_{dw}$ for node $i$ \\
\hline
$E_s$ & Spatial embedding \\
\hline
$E_{s}^i$ & $E_s$ for node $i$\\
\hline
$E_{su}$ & Unknown spatial embedding \\
\hline
$E_{su}^i$ & $E_{su}$ for node $i$ \\
\hline
$E_{sp}$ & Predefined graph embedding \\
\hline
$E_{sp}^i$ & $E_{sp}$ for node $i$ \\
\hline
$E_{st}$ & Spatio-temporal embedding generated as the output of MLP module B\\
\hline
$E_{d}$ & Data Embedding \\
\hline
$E$ & The entire embedding generated as the output of MLP module C \\
\hline
$E_{X}$ & Example of MLP module input denoted by "X"\\
\hline
$E_{X}'$ & Example of MLP module output denoted by "X" \\
\hline
$d_X$ & Temporal dimension of $E_{X}$ and $E_{X'}$\\
\hline
$d_{t}$ & Temporal dimension of $E_t$ \\
\hline
$d_{td}$ & Temporal dimension of $B_{td}$ \\
\hline
$d_{dw}$ & Temporal dimension of $B_{dw}$ \\
\hline
$d_s$ & Temporal dimension of $E_s$ \\
\hline
$d_{su}$ & Temporal dimension of $B_{su}$ \\
\hline
$d_{sp}$ & Temporal dimension of $B_{sp}$ \\
\hline
$d_{st}$ & Temporal dimension of $E_{st}$ \\
\hline
$d_{d}$ & Temporal dimension of $E_d$ \\
\hline
$d$ & Temporal dimension of $E$ \\
\hline
$K$ & Count of discrete time intervals in a day \\
\hline
$K'$ & Count of days in a week \\
\hline
$N$ & Number of nodes/sensors \\
\hline
$N'$ & Intended spatial dimension of CM model\\
\hline
$n_A$ & Number of blocks in MLP module A \\
\hline
$n_B$ & Number of blocks in MLP module B \\
\hline
$n_C$ & Number of blocks in MLP module C \\
\hline
$Q$ & Target / Forecast window \\
\hline
$T$ & Context / Look-back window \\
\hline
$T'$ & Intended temporal dimension of CM/CI model\\
\hline
$T_{td}$ & Input sequence with TD information \\
\hline
$T_{dw}$ & Input sequence with DW information \\
\hline
$t_{td}$ & One-hot vector transformed from the last value of $T_{td}$ \\
\hline
$t_{dw}$ & One-hot vector transformed from the last value of $T_{dw}$ \\
\hline
$X$ & Input data \\
\hline
$\hat{X}$ & Output / Forecast result \\
\hline

\end{tabular}
\end{center}
\caption{Notations in ST-MLP.}
\label{Notations}
\end{table}

\subsection{Dataset Description}

Our experiments encompass a traffic speed dataset and three traffic volume datasets, as summarized in Table~\ref{dataset}.

\subsubsection{Traffic Speed Dataset}
PEMS-BAY comprises six months of speed data collected from 325 static detectors situated in the San Francisco South Bay Area.
\subsubsection{Traffic Volumne Dataset}
PEMS04, PEMS07, and PEMS08 are traffic volume flow datasets that provide real-time highway traffic volume information in California. These datasets are sourced from the Caltrans Performance Measurement System (PeMS) and are collected at 30-second intervals. The raw traffic flow data is subsequently aggregated into 5-minute intervals.
\begin{table}[!ht]
\begin{center}
\renewcommand{\arraystretch}{1.5} 
\setlength{\tabcolsep}{5pt}    
\footnotesize

\begin{tabular}{|c|c|c|c|c|}
\hline
Dataset & PEMS-BAY & PEMS04 & PEMS07 & PEMS08 \\ 
\hline
Type & Speed & Volumes & Volumes & Volumes \\ 

\hline
Time Span & 5 months & 2 months & 3 months & 2 months \\ 
\hline
Num of Nodes & 325 & 307 & 883 & 170 \\ 
\hline
Record Steps & 5 mins & 5 mins & 5 mins & 5 mins \\ 
\hline

\end{tabular}
\end{center}
\caption{ Statistics of datasets.}
\label{dataset}
\end{table}

\subsection{Baseline Methods}

We compare the proposed model (ST-MLP) with the following methods for traffic forecasting:

\begin{itemize}
\item[$\bullet$] Historical Inertia (HI)~\cite{cui2021historical}: A baseline method that utilizes the most recent historical data points in the input time series.
\item[$\bullet$] Graph WaveNet~\cite{wu2019graph}: A framework that combines an adaptive adjacency matrix with graph convolution and 1D dilated convolution.
\item[$\bullet$] DCRNN~\cite{li2017diffusion}: A module that integrates graph convolution into an encoder-decoder gated recurrent unit.
\item[$\bullet$] AGCRN~\cite{bai2020adaptive}: Graph convolution recurrent networks incorporating adaptive graphs to capture dynamic spatial features.
\item[$\bullet$] STGCN~\cite{yu2017spatio}: A framework that combines graph convolution and a 1D convolution unit together.
\item[$\bullet$] StemGNN~\cite{cao2020spectral}: A structure that captures inter-series correlations and temporal dependencies jointly in the spectral domain.
\item[$\bullet$] GTS~\cite{shang2021discrete}: A graph neural network structure that extracts unknown graph information.
\item[$\bullet$] MTGNN~\cite{wu2020connecting}: A module that developed based on a WaveNet backbone.
\item[$\bullet$] DGCRN~\cite{li2023dynamic}: A graph convolution recurrent network which generates a dynamic graph at each time step.
\item[$\bullet$] DLinear~\cite{zeng2023transformers}: A simple MLP framework with the CI strategy designed for long-term time series forecasting.
\item[$\bullet$] PatchTST~\cite{nie2022time}: A Transformer-based model with the CI strategy designed for long-term time series forecasting.
\item[$\bullet$] STID~\cite{shao2022spatial}: Spatial-temporal identity, an effective MLP method by attaching spatial and temporal identity information.
\item[$\bullet$] STNorm~\cite{providence2022spatial}: An innovative technique that applies spatial normalization and temporal normalization.
\end{itemize}

\subsection{Implementation Platform}
We train all the models on the AMD EPYC 7642 Processor @ 2.30GHz with a single RTX 3090 (24GB) GPU. We modify the codes from BasicTS \footnote{https://github.com/zezhishao/BasicTS}~\cite{shao2022basicts} and easytorch \footnote{https://github.com/cnstark/easytorch}~\cite{wang2020easytorch} to implement our model and all baselines.



\subsection{Hyperparameters}

The hyperparameters used for reproducing our experiment results are shown in the Table \ref{tab:hyper}.

\begin{table}[!h]
\renewcommand{\arraystretch}{1.1}
\LARGE
\begin{adjustbox}{width=0.47\textwidth}
\begin{tabular}{|c|cccc|}

\hline
Dataset & \multicolumn{1}{c|}{PEMS-BAY} & \multicolumn{1}{c|}{PEMS04} & \multicolumn{1}{c|}{PEMS07} & PEMS08 \\ \hline
batchsize       & \multicolumn{4}{c|}{32}                                                                              \\ \hline
learning\_rate       & \multicolumn{4}{c|}{0.002}                                                                              \\ \hline
weight\_decay      & \multicolumn{4}{c|}{0.0001}                                                                              \\ \hline
gamma       & \multicolumn{4}{c|}{0.5}                                                                              \\ \hline
milestones       & \multicolumn{4}{c|}{[1, 50, 80]}                                                                              \\ \hline
num\_epochs       & \multicolumn{4}{c|}{200}                                                                              \\ \hline
$n_A$       & \multicolumn{4}{c|}{1}                     \\ \hline
$n_B$       & \multicolumn{4}{c|}{1}                     \\ \hline
$n_C$       & \multicolumn{4}{c|}{3}                     \\ \hline
$d_{d}$        & \multicolumn{4}{c|}{96}                                                                              \\ \hline
$d_{su}$       & \multicolumn{4}{c|}{32}                     \\ \hline
$d_{sp}$       & \multicolumn{4}{c|}{32}                     \\ \hline
$d_{td}$       & \multicolumn{4}{c|}{32}                                                                              \\ \hline
$d_{dw}$       & \multicolumn{1}{c|}{32}         & \multicolumn{1}{c|}{32}       & \multicolumn{1}{c|}{16}       &    32    \\ \hline
Normalization       & \multicolumn{2}{c|}{LayerNorm}                                       & \multicolumn{2}{c|}{BatchNorm}                \\ \hline
\end{tabular}
\end{adjustbox}
\caption{Hyperparameters for ST-MLP implementation.}
\label{tab:hyper}
\end{table}

\subsection{Efficiency Study}

In this section, we aim to present the raw training times for each epoch of various STGNN models and ST-MLP across different datasets. Table~\ref{efficiency} clearly illustrates that ST-MLP boasts the shortest training time among them.

\begin{table}[!h]
\footnotesize 
\begin{center}
\renewcommand{\arraystretch}{2} 
\setlength{\tabcolsep}{6 pt}    
\begin{tabular}{|c|c|c|c|c|c|}
\hline
Dataset & PEMS-BAY & PEMS04 & PEMS07 & PEMS08 \\ 
\hline
\hline
Methods & \multicolumn{4}{c|}{Seconds/epoch} \\ 
\hline
\hline
AGCRN & 62.82  & 16.48 & 86.55 & 12.71 \\ 
\hline
DCRNN & 219.12  & 56.93 & 312.73 & 35.78 \\ 
\hline
DGCRN & 215.29 & 55.95 & 314.96 & 35.68 \\ 
\hline
GWNet & 63.13 & 16.46 & 91.23 & 10.35 \\ 
\hline
GTS & 174.94 & 28.07 & 226.28 & 18.81 \\ 
\hline
MTGNN & 44.29  & 17.50 & 57.25 & 15.07 \\ 
\hline
StemGNN & 28.51 & 7.85 & 54.76 & 6.87 \\ 
\hline
STGCN & 61.21 & 15.26 & 85.32 & 7.53 \\ 
\hline
ST-MLP & 18.22 & 5.34  & 19.05 & 4.23 \\ 
\hline
\end{tabular}
\end{center}
\caption{Absolute training time for one epoch in different datasets. ST-MLP achieves the best efficiency.}
\label{efficiency}
\end{table}

\end{document}